\documentclass[conference]{IEEEtran}
\IEEEoverridecommandlockouts
\usepackage{cite}
\usepackage{comment}
\usepackage{amsmath,amssymb,amsfonts}
\usepackage{algorithmic}
\usepackage{graphicx}
\usepackage{textcomp}
\usepackage{url}
\usepackage{tikz}
\usepackage{xcolor}

\DeclareMathOperator{\topk}{topk}

\def\BibTeX{{\rm B\kern-.05em{\sc i\kern-.025em b}\kern-.08em
    T\kern-.1667em\lower.7ex\hbox{E}\kern-.125emX}}
    
\newcommand\copyrighttext{%
  \footnotesize \textcopyright 2021 IEEE. Personal use of this material is permitted.
  Permission from IEEE must be obtained for all other uses, in any current or future
  media, including reprinting/republishing this material for advertising or promotional
  purposes, creating new collective works, for resale or redistribution to servers or
  lists, or reuse of any copyrighted component of this work in other works.}
\newcommand\copyrightnotice{%
\begin{tikzpicture}[remember picture,overlay]
\node[anchor=south,yshift=10pt] at (current page.south) {\fbox{\parbox{\dimexpr\textwidth-\fboxsep-\fboxrule\relax}{\copyrighttext}}};
\end{tikzpicture}%
}    
    
\begin{document}

\title{Learning distant cause and effect using only local and immediate credit assignment}

\author{\IEEEauthorblockN{David Rawlinson}
\IEEEauthorblockA{\textit{Cerenaut}\\
Melbourne, Australia \\
dave@cerenaut.ai}
\and
\IEEEauthorblockN{Abdelrahman Ahmed}
\IEEEauthorblockA{\textit{Cerenaut}\\
Sydney, Australia \\
abdel@cerenaut.ai}
\and
\IEEEauthorblockN{Gideon~Kowadlo}
\IEEEauthorblockA{\textit{Cerenaut}\\
Melbourne, Australia \\
gideon@cerenaut.ai}}

\maketitle
\copyrightnotice

\begin{abstract}
We present a recurrent neural network memory that uses sparse coding to create a combinatoric encoding of sequential inputs. The network is trained using only local and immediate credit assignment. Despite this constraint, results are comparable to networks trained using deep backpropagation or BackProp Through Time (BPTT). With several examples, we show that the network can associate distant cause and effect in a discrete stochastic process, predict partially-observable higher-order sequences, and learn to generate many time-steps of video simulations. Typical memory consumption is 10-30x less than conventional RNNs, such as LSTM, trained by BPTT. One limitation of the memory is generalization to unseen input sequences. We additionally explore this limitation by measuring next-word prediction perplexity on the Penn Treebank dataset.
\end{abstract}


\section{Introduction}
\label{sec:intro}
Researchers have been unable to find a biological equivalent to the deep error backpropagation used widely in artificial neural networks \cite{o1996biologically,luo2017adaptive}. This presents a credit assignment problem: How do biological neurons determine the influence of a synapse on an error that may occur many layers distant, or many steps in the future? The answer to this question is of practical significance because it may yield alternative, superior learning rules. Existing biologically-plausible approaches to the distant credit assignment problem have noticeable performance limitations when compared to \emph{implausible} learning rules \cite{balduzzi2015kickback,bengio2015towards,ernoult2019updates}. In this paper, we hope to overcome some of these limitations while preserving biological plausibility, using the approach of combinatoric sparse coding.

%\IEEEpubidadjcol

% gideon
% I wanted to highlight basically all subsections of the Intro... especially 'Sequence learning' and 'Gated memory cells'.. which can probably be simplified? Thoughts?

\subsection{Sequence learning}
To date, the most successful architectures for sequence learning are Recurrent Neural Networks (RNNs) and auto-regressive, feed-forward networks \cite{oord2016wavenet}. Truncated Back-Propagation Through Time (BPTT) \cite{sutskever2013training} unrolls the state of an RNN over a fixed number of time steps $t$, allowing errors to be assigned to the relevant synaptic weights. The network cannot be trained to exploit causes earlier than $t$ time-steps.

Autoregressive, feed-forward networks utilize a fixed moving window of $t$ most recent inputs, that are processed simultaneously through several layers. Error gradients are back-propagated through the layers. Like truncated BPTT, context more distant than $t$ steps cannot be utilized. Techniques such as dilated convolutions allow $t$ to be large (100 steps or more).

Credit assignment in the above methods is widely believed to be biologically implausible due to backpropagation through time and layers. Recent work based on Predictive Coding has shown that deep-backpropagation through network layers can be closely approximated using only local computations, but the approach cannot be extended to approximate BPTT \cite{millidge2020predictive}. 

Credit assignment over time can also be achieved by forward-propagating derivatives of hidden states and outputs with respect to each synaptic weight. Real-Time Recurrent Learning (RTRL) \cite{williams1995gradient} is an instance of this concept, but is rarely used in practice due to the difficulty of maintaining viable derivatives. A more recent example is \emph{e-prop} (forward propagation of eligibility traces) but this is only shown to work for short-term association \cite{bellec2019eligibility}.

\subsection{Gated memory cells}
%Another problem (particularly with BPTT and RNNs) is vanishing or exploding gradients. 
In most RNNs, such as Elman networks, iterative partial derivative updates are likely to produce vanishing or exploding gradients. Gated memory layers such as Long-Short-Term-Memory (LSTM) \cite{hochreiter1997long} and Gated Recurrent Units (GRUs) \cite{chung2014empirical} mitigate this problem by creating a more robustly differentiable relationship between memory state, gate operations and outputs.

Using gated memory cells, causes and effects may be successfully associated despite separation by hundreds of steps. However, the memory requirements of gated memory layers are much larger than BPTT with ordinary RNNs, due to the requirement to train many gate-control parameters.

Since we focus on the credit assignment problem, our objective is to demonstrate performance comparable or better than RNNs trained by BPTT. Gated memory architectures trained by BPTT likely represent an upper bound on our expectations, especially considering domain-specific optimizations that our model will lack.

\subsection{Practical requirements}
How much context do we actually need for human-like sequence learning performance? Let's take natural language modelling as an example. Chelba et al \cite{chelba2013one} showed that on the Billion Word benchmark, an LSTM $n$-gram model where $n=13$ was as good as LSTMs with access to longer history. In fact, even with simple smoothed $n$-gram language models such as Kneser-Ney \cite{ney1994structuring}, longer histories do not yield practical advantage, in part because the additional higher-order context is unlikely to generalize well, especially in smaller datasets.

Recent work by Bai et al \cite{bai2018empirical} showed that in general, feed-forward autoregressive models with dilated convolutions tend to outperform RNNs. They also showed that even where very long term context was definitely necessary, RNNs could not exploit it. This implies that the encoding of context is at least as important as the amount of context. 

We draw two conclusions. First, it is essential to verify model performance on real tasks to verify effective access to earlier context. Second, for competitive performance on natural language benchmarks it is likely only necessary to perform credit assignment over a relatively modest context of, perhaps, 100 steps. Since natural languages are tailored to human capabilities, this may be indicative of general human limits. Psychological studies of human problem-solving suggest that we work with relatively simple mental simulations of short sequences of highly abstracted features - longer sequences are made shorter by abstraction \cite{khemlani2013kinematic}.

\subsection{Biologically plausible criteria}
We adopted the following criteria for biological plausibility:

\begin{itemize}
  \item Only local credit assignment. No back-propagation of errors between cell-layers
  \item No synaptic memory beyond the current and/or next step
  \item No time-travel, making use of past or future inputs or earlier hidden states
\end{itemize}

Note that we only aim to address the distant credit assignment problem, and do not claim that our approach is otherwise an accurate reproduction of biological network learning.  

The computational capabilities of single-layer networks are very limited, especially in comparison to two-layer networks. Biological neurons perform ``dendrite computation'', involving integration and nonlinearities within dendrite subtrees \cite{guerguiev2017towards}. This is computationally equivalent to 2 or 3 typical artificial neural network layers. For this reason we allow ourselves to use error backpropagation across two ANN layers, under the assumption that this could approximate dendrite computation within a single biological cell layer, and training signals inside cells.

\subsection{Memorization approach}
Backpropagation through time enables networks to selectively access hidden states and input from many steps prior. If we can only back-propagate errors across 2 or 3 time steps, how can we learn to retain data that won't be useful for tens or hundreds of steps? 

One option is to simply remember everything, which means we could use any past observation to predict a future effect. At first glance this seems impractical, even if we were to limit observations to one hundred steps; but Hawkins \& Ahmad \cite{hawkins2016neurons} showed that a form of sparse coding can allow networks to have vast representational capacity, by virtue of all the \emph{combinations} in which cells can be active. 

Sparse coding simply means that most cells have zero value - they are inactive \cite{olshausen1997sparse}. In Hawkins \& Ahmad's HTM model, the few active cells jointly represent not just the current input, but previous inputs as well. This is achieved by applying local inhibition to groups of cells - within each group, different cells fire in response to different sequential contexts. Since a long context is encoded in the current state without filtering for salience, we will refer to this strategy as the memorization approach.

\section{Method}
\label{sec:method}
We will attempt to implement the memorization approach to sparse sequence encoding using a more conventional RNN, trained with stochastic gradient descent. For convenience we will refer to our approach as recurrent sparse memory (RSM). RSM is a predictive autoencoder derived from the sparse autoencoders developed by Makhzani and Frey \cite{makhzani2013k,Makhzani2015}. In their algorithm, a fixed top-$k$ sparsity is applied; sparseness is the only nonlinearity. 

\begin{figure}  
  \centering
  \includegraphics[width=0.95\columnwidth]{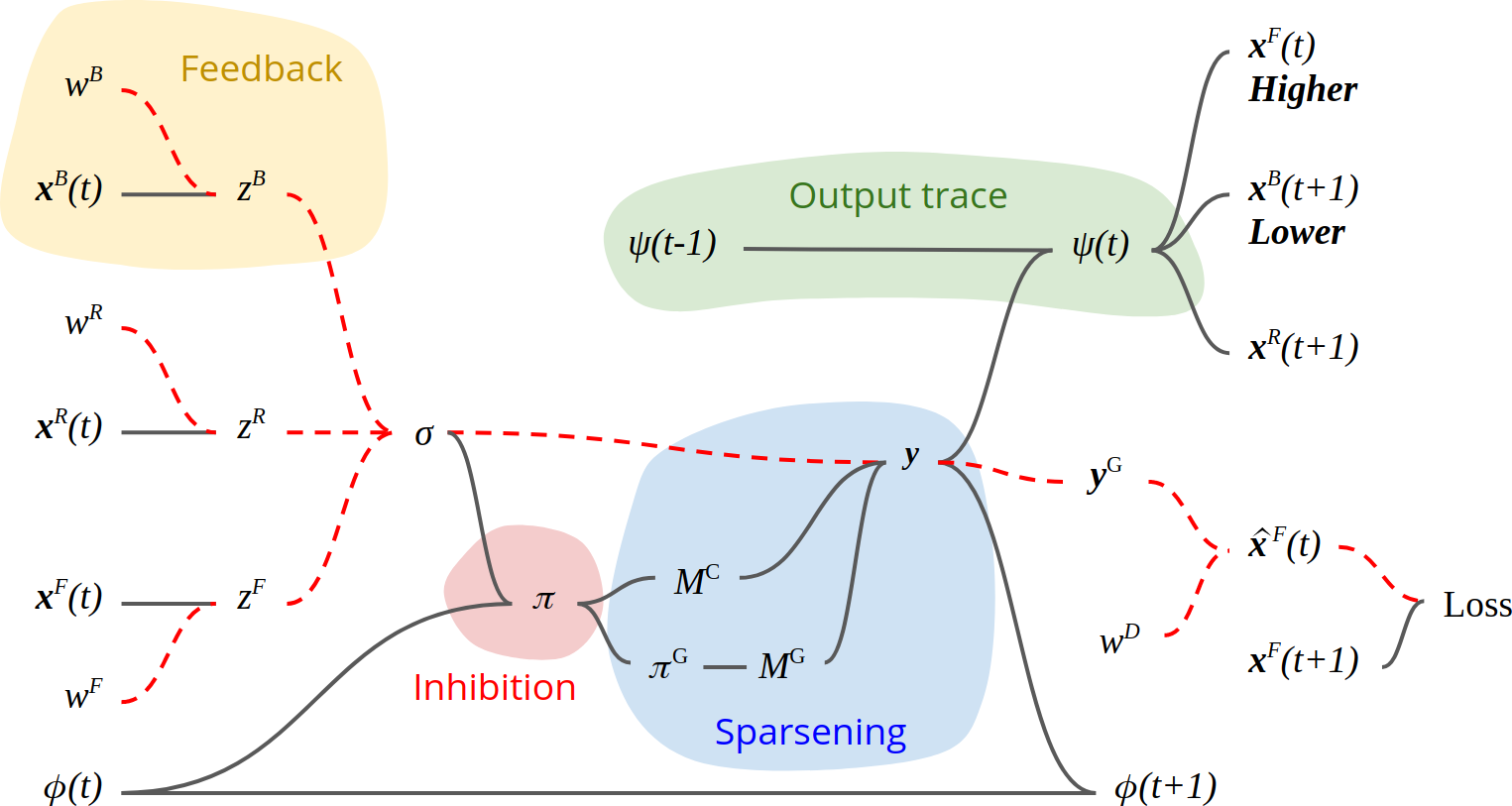}
  \caption{RSM diagram. RSM is trained to generate predictions $\mathbf{\hat{x}}^F(t)$ of its next external input $\mathbf{x}^F(t+1)$. Dashed red lines show how this loss propagates to the learned parameters $w^F$, $w^R$, $w^B$ and $w^D$. The single-layer model is a standard recurrent network with the addition of inhibition (red patch), rank-based sparse masking (blue patch) and integration \& normalization of recurrent input (green patch). The feedback input and parameters are only required if stacking more than one RSM layer with bi-directional connections.}
\label{fig:5}
\end{figure}

An autoencoder learns to reconstruct the current external input $\mathbf{x}^F(t)$ through a bottleneck. To make an autoencoder learn transitions, we instead train it to generate a prediction $\mathbf{\hat{x}}^F(t)$ of the next external input $\mathbf{x}^F(t+1)$ using a mean square error loss.

To allow the network to use previous inputs as context for the prediction, we add a recurrent input $\mathbf{x}^R$, containing a copy of layer cell activity from the previous step - similar to an Elman RNN architecture.

In some experiments we use more than one RSM layer. In this case we require an additional input from the adjacent higher layer $\mathbf{x}^B$. The superscript identifiers $F,R,B$ stand for feed-Forward, Recurrent, and feed-Back respectively.

Cells within the layer are organized into groups. All cells in a group share the same learned weights for external input $\mathbf{x}^F$. Each cell has its own set of fully-connected learned weights for recurrent input $\mathbf{x}^R$. The parameter $g$ specifies the number of groups. Parameter $c$ is the number of cells in each group. 

\subsubsection{Theory of operation}
The network described above is similar to a conventional recurrent layer. However, we add an inhibition term $\phi$ (defined below) that causes cells to become selectively active in particular sequential contexts. The same observation in different contexts is then represented by a different set of active cells. 

When a group of cells responds strongly to an input $\mathbf{x}^F$, the least inhibited cell in the group becomes active. Inhibition due to the refractory period ensures that all cells in a group learn unique contexts in which to fire. Fixed sparsity ensures that cells and groups learn to be active in a huge variety of unique combinations. Each combination of cells encodes not just the current input, but a long history of prior inputs as well. As in \cite{hawkins2016neurons}, the memory can represent many long histories using the current set of active cells, in effect learning a higher-order representation of these sequences.

% gideon
% I think the 'basic' equations... are surpsingly taking up a lot of room here
% how useful are they? e.g. eqs 1,2,3,7,8

\subsubsection{Model Details}
See Figure \ref{fig:5} for an overview of the model. We can calculate the matrix product of inputs and weights as normal for a single ANN layer. Note that each cell has a unique set of recurrent weights $w^R$, whereas all cells in a group share the same external input weights $w^F$:

\begin{equation}
\mathbf{z}^F = w^F\mathbf{x}^F(t)
\end{equation}
\begin{equation}
z^R = w^R\mathbf{x}^R(t)
\end{equation}
\begin{equation}
z^B = w^B\mathbf{x}^B(t)
\end{equation}

The weighted sum $\sigma_{ij}$ of a cell $j$ in group $i$ is given below. $\sigma$ is a matrix of dimension (groups $\times $ cells) i.e. $g \times c$:

\begin{equation}
\sigma_{ij} = \mathbf{z}^F_i + z^R_{ij} + z^B_{ij}
\end{equation}

%This implements the notion that cells in a group share an external input dendrite with weights $w^A$, but have individual recurrent input dendrites with weights $w^B$.

We shift the weighted sums to positive nonzero values and calculate cell activity $\pi$ after applying an inhibition term $\phi$: 

\begin{equation}
\pi_{ij} = (1-\phi_{ij}(t)) \cdot (\sigma_{ij} - \min(\sigma) +1)
\end{equation}

We reduce cell activity $\pi$ to group activity $\pi^G$ by taking the max value of the cells in each group:

% https://math.stackexchange.com/questions/40861/mathematical-notation-for-the-maximum-of-a-set-of-function-values

\begin{equation}
\pi^G_{i} = \max( \pi_{i1}, \dotso, \pi_{in} )
\end{equation}

The next step is to calculate two sparse binary masks $M$. $M^C$ indicates the most active cell in each group. $M^G$ indicates the most active groups in the layer. We use a function $\topk(a,b)$ that returns a `1' for the top $b$ elements in the last dimension of argument $a$, and `0' otherwise.

\begin{equation}
M^C = \topk(\pi,1)
\end{equation}
\begin{equation}
M^G = \topk(\pi^G,k)
\end{equation}

We take the elementwise product of the weighted sum and the two masks to yield the weighted sum of one cell from each of the top-$k$ groups. This is the sparsening step. A nonlinearity is applied.

\begin{equation}
\mathbf{y}_{ij} = \tanh( \sigma_{ij} \cdot M^G_{i} \cdot M^C_{ij} )
\end{equation}

Cells that are selected are inhibited in future, mimicking the refractory period observed in biological neurons. The refractory period also ensures good utilization of all cells during training. Inhibition decays exponentially, but since we use ranking to select groups, even tiny inhibitions can have a significant effect over long periods. The period of effective inhibition depends on the number of competing groups and cells available rather than the rate of decay. The hyperparameter $0 \leq \gamma \leq 1$ determines the inhibition decay rate.

\begin{equation}
\phi_{ij}(t+1) = \max( \phi_{ij}(t) \cdot \gamma , \mathbf{y}_{ij} )
\end{equation}

Finally, we allow the option to integrate the recurrent input over time. $\epsilon$ determines the decay rate of the integrated encoding $\psi$. Since the current value of $\psi$ theoretically represents all previous states, when external input statistics are identical between training and test time, $\epsilon$ can be zero. However, we found a generalization advantage to nonzero $\epsilon$, discussed later.

\begin{equation}
\mathbf{\psi}(t) = \max( \mathbf{\psi}(t-1) \cdot \epsilon , \mathbf{y} )
\end{equation}

$\mathbf{x}^R$ is both the recurrent input and input to any task-specific classifier (see experiments, below). If more than one RSM layer is used, $\mathbf{x}^B_{lower} = \mathbf{x}^R_{higher}$ and $\mathbf{x}^F_{higher} = \mathbf{x}^R_{lower}$.
$\alpha$ is a normalizing scalar such that the sum of $\mathbf{x}^B$ is 1 (since we have many zero values, and few or no negative values, we have not found a more appropriate norm). $\mathbf{x}^R$ is initialized as zeros and updated:

\begin{equation}
\mathbf{x}^R(t+1) = \alpha \cdot \mathbf{\psi}(t)
\end{equation}

Each RSM layer is trained to predict the next input $\mathbf{x}^F(t+1)$. Prediction $\mathbf{\hat{x}}^F$ is generated by ``decoding'' through a bottleneck of masked hidden layer group activities $\mathbf{y}^G$ by taking the max activity of the cells in each group $i$:

\begin{equation}
\mathbf{y}^G_{i} = \max( \mathbf{y}_{i1}, \dotso, \mathbf{y}_{in} )
\end{equation}

$w^D$ is a set of decoding weights of dimension equal to the transpose of $w^F$. We observed that tied weights $w^F = (w^D)^T$ were less effective in our experiments.

\begin{equation}
\mathbf{\hat{x}}^F(t) = w^D\mathbf{y}^G
\end{equation}

The mean square error between prediction $\mathbf{\hat{x}}^F(t)$ and observation $\mathbf{x}^F(t+1)$ is minimized by stochastic gradient descent. Gradients propagate from the prediction through $w^D$ to the sparse bottleneck $\mathbf{y}^G$, and subsequently to the encoding weights $w^F$, $w^R$ and $w^B$. The latter two learn to modulate hidden activity given the sequential context. The backpropagation depth is fixed by this architecture to 2. We conceptualize an RSM layer as a single layer of neurons with two or three independently integrated dendrite trees. Max and ranking functions respectively represent local and regional competition between cells.

The architecture is extremely memory efficient, requiring only a single copy of prior memory state $\mathbf{x}^R$ and inhibition $\phi$ in addition to current external input $\mathbf{x}^F$ for both training and encoding. If $c$ is the number of cells, the asymptotic measure of RSM memory use is $O(c)$. Both truncated BPTT and feed-forward autoregressive approaches such as WaveNet \cite{oord2016wavenet} require $O(ct)$ where $t$ is the time horizon. With the exception of top-k ranking, all the functions described above are fast, simple arithmetic operations.

%Within the exception of ranking, all computations are simple, fast operations.

\subsection{Experimental Architecture}
\begin{figure}
\centering
    \includegraphics[width=0.95\columnwidth]{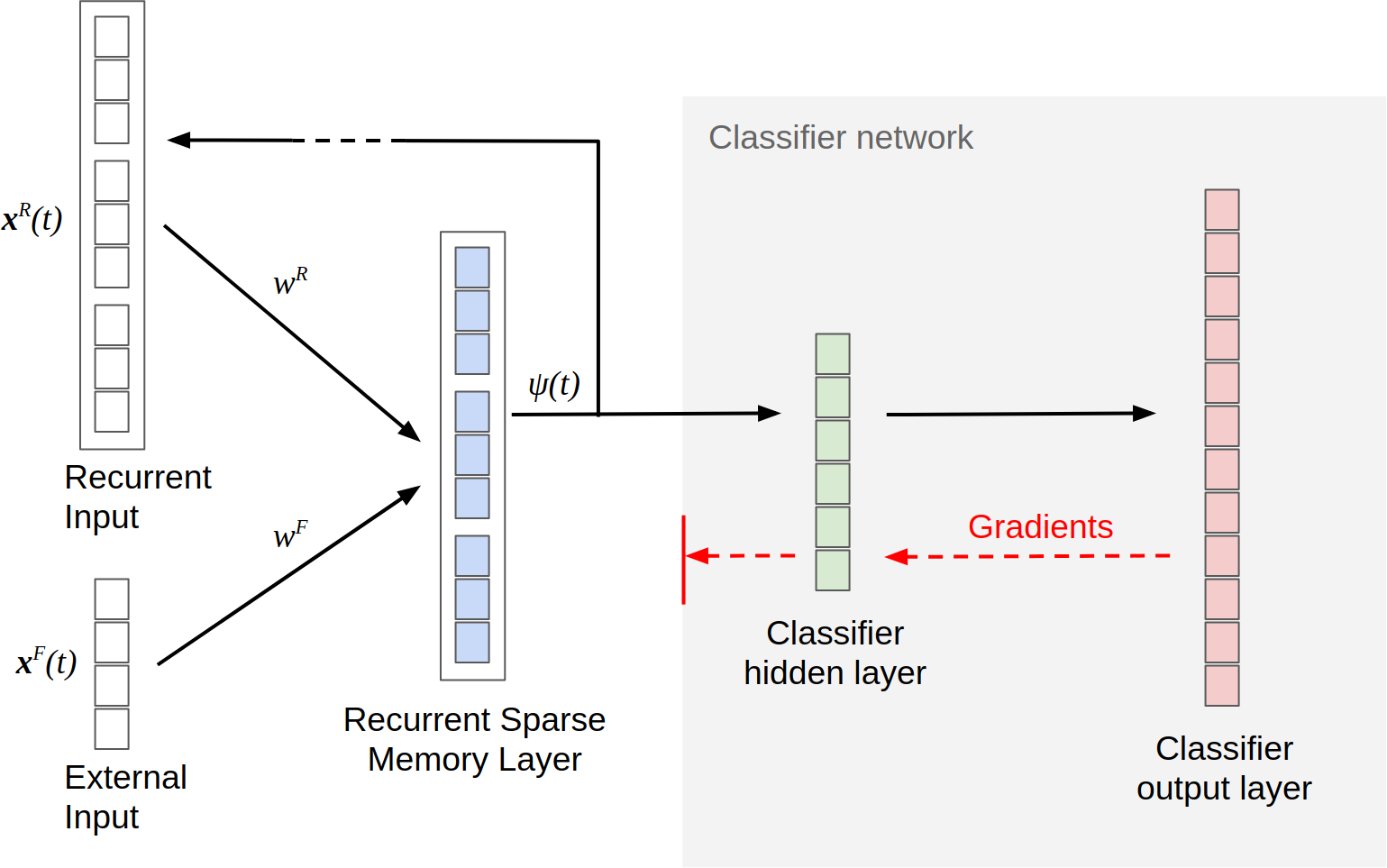}
  \caption{The architecture used in all experiments, except the multi-step prediction task where the classifier network is replaced with a GAN. The classifier has 2 trainable layers, hidden and output. Gradients only propagate within the RSM layer, and within the two classifier layers. The maximum backpropagation depth is 2.}
\label{fig:1}
\end{figure}

For most of our experiments we use a single RSM layer. To exploit the memory for a discriminatory purpose, we add a ``classifier'' network of two fully-connected layers (see Figure \ref{fig:1}). We use leaky-ReLU nonlinearities in these networks to reduce the accumulation of dead cells due to nonstationary RSM encoding. As per our rules, gradients do not propagate from the classifier into the RSM layer, or across the recurrent or feedback inputs. All layers are trained simultaneously \& continuously using only the current input, satisfying the objective of learning using local and immediate credit assignment.

With the addition of the classifier, the architecture is reminiscent of ``Reservoir Computing'' models \cite{lukovsevivcius2009reservoir}, with the primary differences being the memory is trained via stochastic gradient descent (reservoir ``memory'' is typically not trained), having sparse activation to improve the encoding properties, and inhibition to recruit larger cell populations during training.

\section{Experiments}
\label{sec:experiments}
All experiments use the same architecture described above, with the exception that the classifier is replaced with a Generative Adversarial Network (GAN) \cite{goodfellowgan2014} in the generative multi-step prediction task.\footnote{The codebase and a full list of hyperparameters for each experiment are publicly available on GitHub at: \url{https://github.com/Cerenaut/rsm}} %All components of the architecture were trained simultaneously and continuously.% for added biological plausibility.

Published results concerning a sparse memorization approach such as HTM \cite{Cui2016} focus mostly on network robustness and do not enable easy comparison to more popular ANNs. We will attempt to validate the approach on a range of more easily contextualized tasks \& benchmarks.

\subsection{Associating distant cause and effect}
LSTM was specifically designed to enable learning of distant cause and effect, using gated memory cells. %At that time, BPTT and Real-Time Recurrent Learning \cite{rtrl} were dominant ANN approaches to modelling of cause and effect in sequences. 
%LSTM effectively solved the gradient problem using gated memory cells. 
Hochreiter and Schmidhuber demonstrated LSTM using several example problems, including the Embedded Reber Grammar (ERG) \cite{hochreiter1997long}. Since they showed that ordinary RNNs were unable to solve the task, it is a good benchmark for demonstrating distant cause \& effect learning.

\begin{figure}  
  \centering
    \includegraphics[width=0.6\columnwidth]{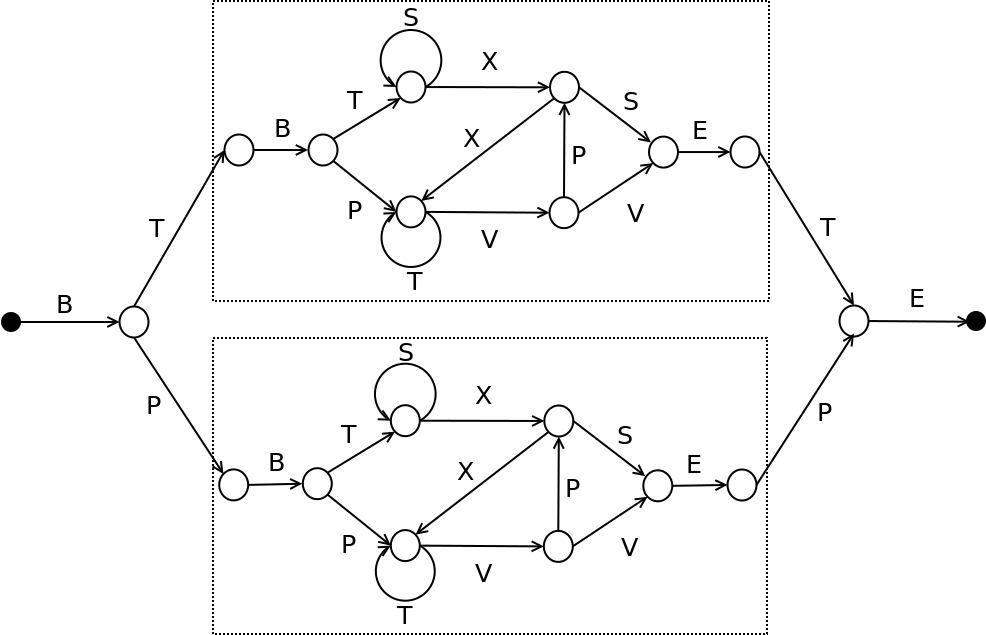}
  \caption{The Embedded Reber Grammar problem. A graph is used to generate sequences of observations, denoted by letters annotating the edges of the graph. The graph begins with a fork: B-T or B-P. The two fork paths have identical embedded `Reber' grammars (dashed boxes). After a Reber grammar, T or P must be predicted correctly. The Reber grammar generates long, random, distraction sequences. Successful prediction requires the original fork to be remembered. RSM enables the attached classifier to achieve high accuracy without any other context, despite long, random distraction sequences.}\label{fig:erg}
\end{figure}

The ERG has an initial fork (T,P), followed by the same distraction subsequence on both forks (see Figure \ref{fig:erg}). The final step (T,P) can only be predicted by remembering the pre-fork symbol. For added difficulty, symbols T and P also occur in the distraction sequence. 

RSM enables its attached classifier to achieve over $99\%$ accuracy on a random sample from the grammar (see table \ref{table:erg}).

The ERG is not deterministic. It generates sequences of minimum length 9. The maximum length is unbounded, but to achieve 99\% accuracy, all sequences of length $\leq 30$ must be predicted correctly. Without gated memory cells, simple RNNs trained with BPTT and RTRL fail to solve the ERG task \cite{hochreiter1997long}. Cui et al report that HTM achieved 98\% accuracy \cite{Cui2016}.

% RTRL https://www.dlsi.ua.es//~mlf/nnafmc/pbook/node29.html

% gideon
% all table captions are QUITE DENSE... maybe this looked better in the previous template.. but doesn't look that great here.
% applies to all tables

\begin{table}[htbp]
\caption{Distant cause \& effect prediction accuracy on the Embedded Reber Grammar task for various neural models}
%\caption{Distant cause \& effect prediction accuracy on the Embedded Reber Grammar task for various neural models. The LSTM architecture was the first to solve this task. A conventional (Elman) RNN trained by BPTT cannot solve this task \cite{hochreiter1997long}. The conceptually similar HTM and RSM networks both provide an approximate solution.}
\begin{center}
\begin{tabular}{|c|c|c|c|}
\hline
\textbf{Model} & \textbf{Accuracy}  \\
\hline
Elman RNN \cite{hochreiter1997long} & 0 \\
\hline
LSTM \cite{hochreiter1997long} & 100 \\
\hline
HTM \cite{Cui2016} & 98.4 \\
\hline
\textbf{RSM} & \textbf{99.2} \\
\hline
\end{tabular}
\end{center}
\label{table:erg}
\end{table}

We speculate that RSM achieves high accuracy by memorizing all the sequences frequently generated by the grammar. In a sample of 5000 sequences drawn from the grammar, we observed 601 unique sequences. We used $g=200$, $c=6$ and $k=25$ for this experiment. Since there are only 48 disjoint sets of 25 cells, memorization of 600 or more sequences implies that RSM has learned a combinatoric model.

RSM prediction is not confounded by stochastic sequences, but note that the statistics of training and test sequences are identical. Both the RSM layer and classifier components of the network achieve the same accuracy, but the classifier allows a specific label to be predicted, rather than an image of a label.

\subsection{Modelling partially observable, higher-order sequences}
We have demonstrated that RSM is able to model non-deterministic sequences, but what about sequences that are only partially observable? Since an RSM layer is easily derived from a convolutional network layer, and can handle non-deterministic sequences, it is likely RSM is able to simultaneously learn sequential and spatial structure in its input. To test this hypothesis, we presented repeating sequences of MNIST images and measured next-digit appearance and label prediction accuracy. Each step in a sequence has a specific label, but a random image of that label is selected to represent it as input. Due to variation in digit images, the actual sequence is only partially observable. True labels are never observed by the RSM. There are no sequence boundaries and the network is never reset to a baseline state, so all structure must be inferred from the stream of input.

\begin{table}[htbp]
\caption{MNIST image sequence prediction experiment}
%\caption{MNIST image sequence prediction experiment. This task combines learning of spatial structure and higher-order sequences. RSM is trained to predict the next input given prior inputs, for sequences with varying Markov order. While the sequences are deterministic, each observation provided as input is a random MNIST image of the given number. RSM output is therefore a prediction of the next digit-image. RSM does not have access to the actual numbers during training or testing. The attached classifier network is trained with one-hot class vectors representing the numbers, but does not have access to any context, only RSM hidden state. For reference, we also include prediction of the current image label without context (i.e. a conventional image classification task).}
\begin{center}
\begin{tabular}{|c|c|c|c|}
\hline
\textbf{Sequence} & \textbf{Markov order} & \textbf{Accuracy}  \\
\hline
`0,1,2,3,4,5,6,7,8,9' & 1 & 99.9 \\
\hline
`0,1,2,3,4, 0,4,3,2,1' & 2 & 99.9 \\
\hline
`0,1,2,3, 0,1,2,3, 0,3,2,1' & 5 & 99.9 \\
\hline
None (image classification) & 0 & \textbf{98.2} \\
\hline
\end{tabular}
\end{center}
\label{table:pos}
\end{table}

Table \ref{table:pos} shows some of the sequences tested and the minimum history necessary for perfect prediction. Note that some sequences have Markov order $>1$, meaning that correct recognition of the current image is insufficient to predict the next label. 

\begin{figure}
  \centering
    \includegraphics[width=0.6\columnwidth]{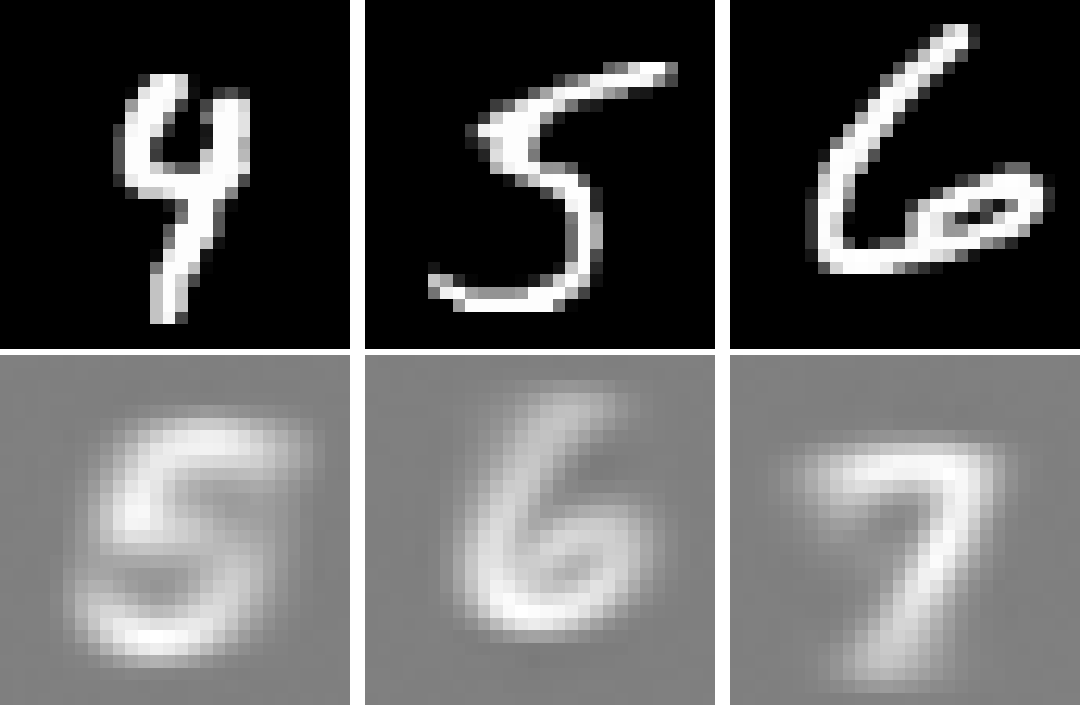}
\caption{MNIST image sequence prediction experiment. This task combines learning of spatial structure and higher-order sequences. Three separate examples are shown in three columns. Top row shows current RSM input images $x^F(t)$. Bottom row shows predicted images $\hat{x}^F(t)$ given by RSM, for a sequence of the numbers 0,1,...,9 repeating. The predictions shown here are correct (i.e. 4 predicts 5, 5 predicts 6 etc.)}\label{fig:3}
%  \caption{MNIST image sequence prediction experiment. This task combines learning of spatial structure and higher-order sequences. Three separate examples are shown in three columns. Top row shows current RSM input images $x^F(t)$. Bottom row shows predicted images $\hat{x}^F(t)$ given by RSM, for a sequence of the numbers 0,1,...,9 repeating. The predictions shown here are correct (i.e. 4 predicts 5, 5 predicts 6 etc.) Random exemplar images are selected as model input for each instance of a number, introducing observational uncertainty - RSM does not have access to the actual numbers during training or testing. Note that despite the uncertain appearance of each digit, RSM is able to generate a correct ``generic'' digit prediction. The add-on classifier gives the correct label prediction with $> 99\%$ accuracy. Although this sequence has a Markov Order of 1, higher order sequences were also tested successfully.}\label{fig:3}
\end{figure}

As before, both RSM and classifier components achieve accuracy of 99\% or better, but the classifier allows labels to be obtained for quantitative measurement. Examination of the next-image predictions is interesting - the RSM generates generic digit-image predictions (see Figure \ref{fig:3}).

When trained to classify the label of the current image, the same fully-connected classifier network achieves a lower accuracy of 98\%. We conclude that the RSM is able to combine spatial and higher-order sequential information in a single layer to boost image classification accuracy to over 99\% for all sequences tested. In many network architectures incorporating gated memory cells, a convolutional stack is used for spatial dimensionality reduction before consideration of prior state held in memory cells (for example, Srivastava et al \cite{srivastava2015unsupervised} use a pretrained stack to generate ``percepts'' as input to an LSTM layer). Integrated modelling of spatial and temporal data in a single layer is likely advantageous, and has been investigated \cite{byeon2018contextvp}, but this line of research is hampered by the complexity of gated memory layers.

A recent publication by Gordon et al \cite{gordon2019learning} applied a modified version of RSM to a similar task with the additional difficulty of non-deterministic sequences. In this case the modified bRSM was observed to learn the sequence structure better than an LSTM with the same number of trainable parameters.

\subsection{Multi-step video prediction}
% Config
% https://github.com/ProjectAGI/rsm/blob/master/definitions/bouncing-balls/train-gan.json
% Abdels blog post
% https://docs.google.com/document/d/1OFXI1syTlSl9m1SBK4K_F7NjWAydiaCPxlai2Ri2Pqg/edit
% Abdel architecture writeup
% https://docs.google.com/document/d/1eL17lYrmMx7vJsIrt_6OmSGkS0H6cERbQTKMYjHt2iI/edit
% https://drive.google.com/drive/u/1/folders/16HQ0CSy-FdD_weKxFx2XBVgffhUiAdhE
% Video on YouTube: https://youtu.be/-lBFW1gbokg
We wished to explore whether RSM can learn temporal dynamics well enough to generate novel sequences, rather than simply next-step predictions. Models trained to perform single-step prediction tend to perform poorly when used for multi-step, self-looped prediction because they lack a persistent representation of the whole production, such as a sentence \cite{bowman2015generating}. In addition, we wanted to explore a task with more demanding spatial characteristics rather than a single character centered in the image. Multi-step video prediction is an obvious candidate task and current models are unsatisfactory \cite{cenzato2019difficulty}.

\begin{figure}
  \centering
    \includegraphics[width=0.95\columnwidth]{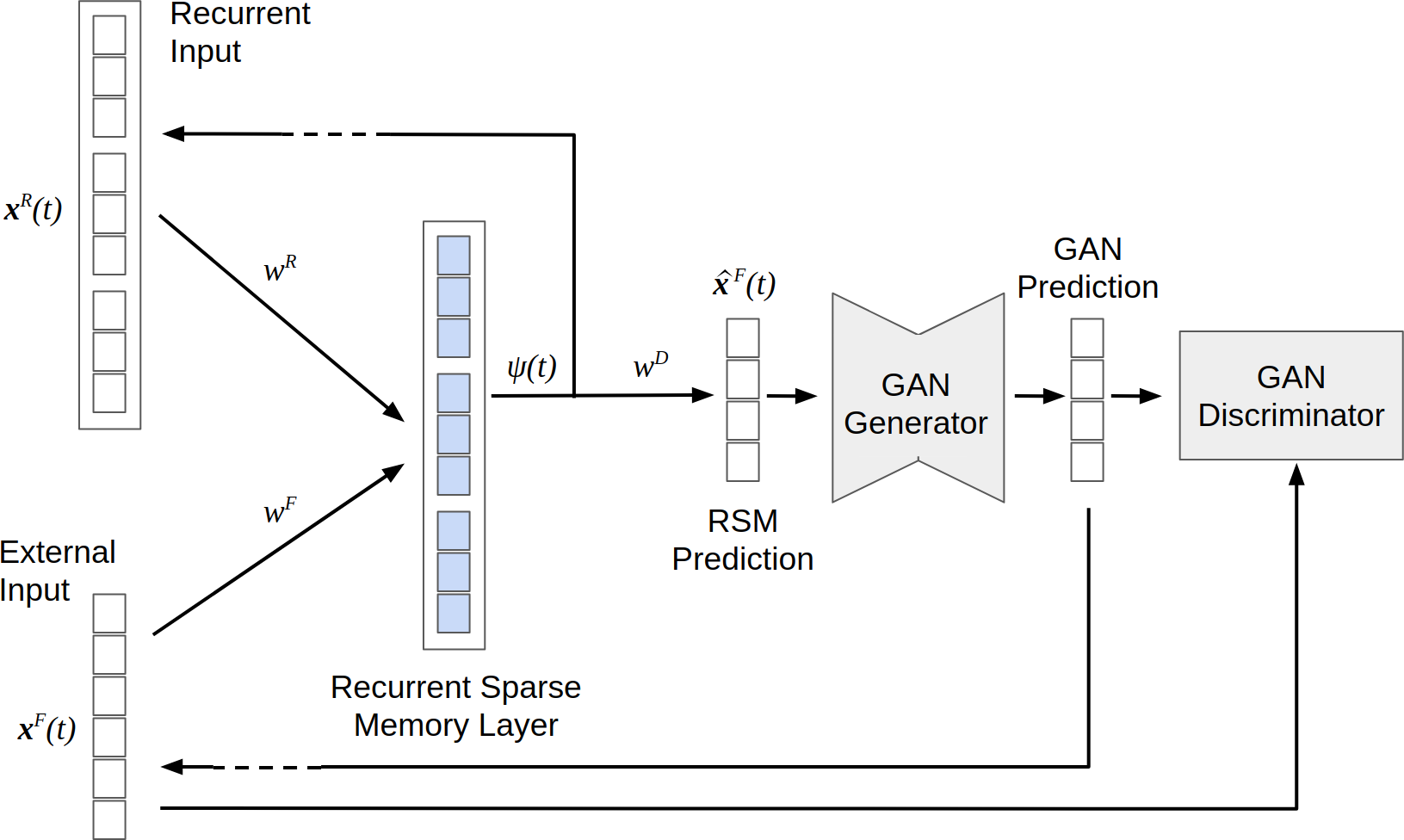}
  \caption{RSM combined with a Generative Adversarial Network (GAN) \cite{goodfellowgan2014}. Although two RSM layers were used, only the lower layer is connected to the GAN and shown here. The upper RSM layer influences the lower RSM layer via the feedback dendrite (not shown). When used in self-looping (generative) mode, RSM predictions approximate the expected value of each pixel, resulting in a slightly blurred appearance that covers all possible ball trajectories (there is some uncertainty in the simulation due to the precise timing of ball and wall interactions). The GAN has no predictive ability, but acts as a rectifier, sharpening RSM predictions into a single specific possible future outcome. Passing the GAN-rectified prediction back into RSM produces good simulations many steps into the future.}\label{fig:rsm_gan}
\end{figure}

\begin{figure}
  \centering
    \includegraphics[width=0.95\columnwidth]{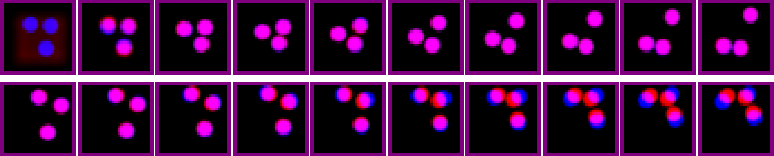}
  \caption{Frame by frame RSM predictions of bouncing ball dynamics (red channel) compared to ground-truth simulation (blue channel). The top row shows RSM in predictive mode, from the start of a sequence. Prediction errors are imperceptible after the 5th step. Bottom row: RSM in generative mode - although the dynamics remain visually realistic for indefinite periods, the trajectories do diverge from the simulation over time.}\label{fig:bball}
\end{figure}

% gideon
% whole paragraph to explain bouncing balls.. does it need this level of detail as it's already been published elsewhere and relatively known benchmark?

The ``Bouncing balls'' video prediction task was introduced by Sutskever et al \cite{sutskever2009recurrent}. The objective is to predict the motion of three billiard balls bouncing around in a box. Interactions between balls and the walls of the box are not precisely predictable from the raster image observations provided as input. Researchers evaluate models trained on this dataset in two ways - first, by measuring next-frame prediction mean-square-error (MSE). Second, qualitatively by generating videos of the learned model in self-looped mode. In this condition, a few ``priming'' frames are produced by the simulation, and passed as input to the learned model. The learned model continuously generates predictions of the next input frame. After priming, the model's own prediction is provided as input. Since tiny angular errors compound rapidly over time, it is expected that sequences generated in this way diverge from the simulation, making an MSE comparison inappropriate (Figure \ref{fig:bball}).

\begin{table}[htbp]
\caption{Next-frame prediction error for the bouncing-balls task}
%\caption{Next-frame prediction error for the bouncing-balls task. RSM is compared to previous results from the [Structured] Recurrent Temporal Restricted Boltzmann Machine ([S]RTRBM) \cite{sutskever2009recurrent,mittelman2014structured}, Deep Temporal Sigmoid Belief Network (DTSBN) \cite{gan2015deep}, Predictive Generative Network (PGN) \cite{lotter2015unsupervised} and a variety of LSTM architectures \cite{cenzato2019difficulty}. In combination with a GAN, RSM provides state of the art next-frame prediction error. Ablation results for RSM single-layer and without GAN are included. These show that a second RSM layer improves the output.}
\begin{center}
\begin{tabular}{|c|c|c|c|}
\hline
\textbf{Model} & \textbf{Error} \\
\hline
Previous frame \cite{lotter2015unsupervised} & $11.82 \pm 0.27$ \\
\hline
RTRBM \cite{sutskever2009recurrent,mittelman2014structured} & $3.88 \pm 0.33$ \\
\hline
SRTRBM \cite{mittelman2014structured} & $3.31 \pm 0.33$ \\
\hline
DTSBN \cite{gan2015deep} & $2.79 \pm 0.39$ \\
\hline
PGN \cite{lotter2015unsupervised} & $0.65 \pm 0.11$ \\
\hline
LSTM \cite{cenzato2019difficulty} & $111.09 \pm 0.68$ \\
\hline
ConvLSTM \cite{cenzato2019difficulty} & $0.58 \pm 0.16$ \\
\hline
Seq2seq ConvLSTM \cite{cenzato2019difficulty} & $1.34 \pm 0.19$ \\
\hline
Seq2seq ConvLSTM multi-decoder \cite{cenzato2019difficulty} & $4.55 \pm 0.40$ \\
\hline
Conv RSM 1L & $3.51 \pm 0.14$ \\
\hline
Conv RSM 2L & $3.72 \pm 0.29$ \\
\hline
Conv RSM 1L + GAN & $0.8 \pm 0.07$ \\
\hline
\textbf{Conv RSM 2L + GAN} & $\boldsymbol{0.41 \pm 0.05}$ \\
\hline
\end{tabular}
\end{center}
\label{table:generative}
\end{table}

\subsubsection{Stacked RSM networks}
The recurrent input $\mathbf{x}^R$ can also be utilized as $\mathbf{x}^F$ for a deeper RSM layer, allowing layers to be stacked. In the stack, credit assignment is still local within each RSM layer (no gradients propagate recurrently or between layers). Stacking up to 3 RSM layers and using the deepest layer as classifier input did not degrade performance in our image sequence experiments, suggesting that RSM layers can be assembled into a hierarchy. In the bouncing balls task, adding a 2nd RSM layer improved prediction accuracy significantly (see table \ref{table:generative}). Since a single RSM layer can learn higher-order models, there is little advantage to deeper networks unless it enables more efficient dimensionality reduction, or association of multi-modal inputs.

\subsubsection{Convolutional RSM}
Given the higher spatial complexity of the bouncing balls task, a convolutional approach is desirable. A convolutional version of RSM can be easily derived from the fully-connected version given above. Learned parameters are shared between all convolutional filter positions, but inhibition $\phi$ and outputs $\psi$ are unique to each position. The number of filter positions (and hence, output dimension $[h^1,w^1]$ of a lower layer) is determined by the feed-forward convolution operation. In a convolutional stack architecture, the number of filter positions in the higher layer $[h^2,w^2]$ may be dissimilar to the lower layer $[h^1,w^1]$. A mini-batch of convolution produces a matrix of dimension $[b,h,w,g,c]$. These dimensions are respectively, batch-size, conv. height, conv. width, groups and cells. For the feedback input $\mathbf{x}^B$ from higher layers, we set the receptive field width and height to $1\times1$ and interpolate from $[h^2,w^2]$ to $[h^1,w^1]$. This ensures that $[h^1,w^1]$ is equal to the number of filter positions required by the feed-forward convolution operation. A $1\times1$ receptive field is also used for the recurrent input, meaning that each filter position only receives a recurrent input from itself.

\subsubsection{GAN rectifier}
We observe that RSM generates a prediction covering all possible next input states (e.g. Figure \ref{fig:3}). Each pixel prediction approximates the expected value of that pixel. When self-looped, the uncertainty in these predictions compounds over time, resulting in degraded input dissimilar to the training input. We noticed that using rudimentary image processing filters to rectify the predictions significantly improved RSM output in self-looped mode. A Generative Adversarial Network (GAN) is well suited to this role, because given a conditioning input it can produce clear, sharp samples that are indistinguishable from genuine input. We trained a GAN to take the lower-layer RSM prediction $\mathbf{\hat{x}}^F$ as conditioning input and output a next-frame sample. Implicitly, the role of the RSM is to predict the distribution of next frame samples, and the role of the GAN is to render one sample from this distribution. The GAN does not have access to any context to allow it to perform a predictive role. Gradients from the GAN do not propagate into the RSM layers. 

We used a stacked convolutional autoencoder as the GAN generator, and a fully-connected network as GAN discriminator. The GAN was trained with both MSE and adversarial losses, copying the Predictive Generative Network (PGN) of Lotter et al \cite{lotter2015unsupervised}. However, the PGN architecture combines CNN and LSTM architectures to add the predictive aspect. In our case we rely entirely on the RSM layers for prediction.

\subsubsection{Video prediction results}
Table \ref{table:generative} shows comparative performance of RSM and works by other authors. For reference, we include results for RSM without GAN and a single layer RSM + GAN. A two layer RSM + GAN system outperforms all prior works, including a variety of LSTM architectures, in terms of next-frame prediction accuracy. To better understand the generated spatio-temporal dynamics in self-looped mode, we recommend viewing a video of the output \footnote{For a qualitative understanding of the generated ball dynamics, see video at: \url{https://www.youtube.com/watch?v=-lBFW1gbokg}}. Although videos of self-looped output are not available for all works, the results are markedly improved over Sutskever et al (RTRBM / SRTRBM) \cite{sutskever2009recurrent}. Whereas the balls wander, change direction or acceleration and stick to each other in the latter, the RSM videos show relatively good conservation of momentum and plausible interactions between objects. 

\subsection{Language modelling: Next word prediction perplexity}
\subsubsection{Generalization}
RSM takes the approach of remembering and encoding a long history of prior states in the current state. To achieve this, the same input in different sequential contexts generates highly orthogonal encodings. This is in contrast to gated memory layers, where a few features are \emph{selectively} remembered, and different sequential contexts can generate \emph{similar} encodings. Consequently, we anticipate that changes in sequential context between training and test datasets will severely disrupt RSM. In other words, we don't expect it to generalize well to unseen sequences.%when test statistics differ from training statistics.

To explore this aspect of generalization, we selected a language modelling task. The Penn TreeBank (PTB) dataset is a corpus of approximately 1 million training words and 80,000 test words. We used Mikolov's preprocessing \cite{mikolov2011empirical}, which results in a dictionary of 10,000 unique words. PTB is known to present a difficult generalization challenge due to the small size of the corpus. Next-word prediction quality is typically measured using perplexity (PPL). Model output is a distribution over the words in the dictionary. In our case, model input is a binary vector representing the (random) indices of each word in the dictionary.

\subsubsection{Generalization features}
For RSM, even small changes between training and test sequences can be disruptive. Imagine that the training corpus contains `The cat sat on the mat' and the test corpus `The cat sat on \emph{a} mat'. Although the RSM encoding would be identical until `on', subsequent states would be highly orthogonal, so the classifier cannot predict `mat' from combinations of active cells never previously observed. To mitigate these types of generalization error, we enabled some minor features.

First, we integrate feedback to the RSM using $\epsilon > 0$. This is not needed to help RSM learn sequences, but it does help RSM \emph{generalize} learned sequences. We apply dropout at rate $0.5$ to encourage the RSM to make use of earlier encodings, for example the active cells representing `The cat sat' are all predictive of a future `mat'. Second, we provide the integrated RSM state to the classifier. These changes reduce test perplexity by 10-15.

The third change is to ``forget'' the recurrent state of the network with a fixed probability $\mu$ during training. If forgetting is randomly selected, we set integrated activity $\psi$ and inhibition $\phi$ to zeros. The intention is to expose the classifier to different subsequences, in effect augmenting the training set. This feature improves test perplexity by another 5-10 points.

\subsubsection{Regularization}
We attempted to regularize the RSM and classifier networks via conventional techniques - in particular adding an L2 loss term, and dropout. We found neither helpful when added to the RSM, but a small L2 value improved classifier performance.

To understand why these techniques are ineffective with this architecture, consider Cui et al \cite{Cui2016}, which was one of the inspirations for our work. They show that sparse distributed representations are extremely robust to network damage, because the meaning of individual elements is correlated. Therefore, to successfully ``ignore'' an observation, many cells' activity must be simultaneously excluded. This is difficult for reasonable levels of both dropout and weight penalties. So it seems ineffective regularization may be a consequence of distributed representations that are robust to damage.

\subsubsection{Model interpolation}
%Smoothing is an essential part of a good language model. 
Mikolov et al \cite{mikolov2011empirical} showed that linear interpolation of model distributions to produce ensemble predictions is very often effective at reducing test perplexity. In particular, earlier RNN and LSTM language models (which we are using as a baseline) were often presented as ensembles. 5-gram language models with Kneser-Ney smoothing (KN5) \cite{chen1999empirical} are optimal for the PTB corpus; higher n-gram models do not deliver better test perplexity. 

When a word occurs, it is likely to occur again in the near future. This general principle leads to cache language models. Mikolov et al\cite{mikolov2011empirical} also showed that caches invariably improve ensemble perplexity. %We decided that a simple unigram cache with exponentially-decaying probability was a biologically plausible feature as it does not violate our rules. 
We felt that adding a fixed, exponentially-decaying probability mass to words after observation - a primitive cache - meets our biological plausibility criteria. We did not allow `adaptive' modelling (i.e. allowing the RSM to learn during the test). We report RSM results interpolated with the simplistic cache, and optionally with KN5 for easier comparison and insights on n-gram ensemble complementarity.

%Context Adaptation in Statistical Machine Translation Using Models with
%Exponentially Decaying Cache
%Jorg Tiedemann -- proper cache, cite this, say not bio plausible.

\subsubsection{Results}

%Test PPL
%KN5 156
%RSM 166
%RSM + KN5 131
%Test acc: 20.6\% explain bimodal result
%Training result - PPL 9 Acc 50\%
With $g=600$ groups, $k=20$, $c=6$ cells per group and 4 epochs of training with batch size 300, we obtained 50\% next word prediction accuracy and a perplexity of 9 on the training corpus. Training perplexity and accuracy had not plateaued. This suggests that moderately sized RSM layers can memorize extremely long sequences.

We found test corpus perplexity to be better with $n=8$ cells per group and only 0.25 epochs of training. We observed next word prediction accuracy of 20.6\% on the test corpus. RSM per-word perplexity is bi-modal, showing rapid oscillations between very low and high values. When not combined with KN5, a small uniform mass significantly reduces average RSM test perplexity (reported result without KN5 includes 7\% uniform mass). The complementarity of RSM and KN5 - evidenced by RSM+KN5 interpolated model perplexity - suggests that RSM sometimes captures longer range context, but also fails to generalize many shorter range contexts.

% python experiment.py --experiment_def ./178c_reload_kn5_only.json
% Perplexity 143.410691 Loss mean 4.965712  Samples 82430.000000
% 
% python experiment.py --experiment_def ./178c_reload_rsm_kn5.json
%       "file_mass":0.66,
%       "uniform_mass":0.0,
%       "input_mass":0.0,
%       "layer_mass":0.26,
%       "cache_mass":0.08,
% Perplexity 124.973573 Loss mean 4.828102  Samples 82430.000000

\begin{table}[htbp]
\caption{PTB test-set perplexity of RSM and later bRSM \cite{gordon2019learning} models, with selected other algorithms for context}

%\caption{PTB test-set perplexity of RSM and later bRSM \cite{gordon2019learning} models, with selected other algorithms for context. RSM should be comparable to other early language models, in particular RNNs trained by BPTT. RSM training set perplexity is included to highlight that while distant context can be successfully exploited, \emph{generalization} of this context is deficient: RSM test set perplexity is worse than comparable architectures trained by BPTT. The subsequent bRSM variant \cite{gordon2019learning} fixed this limitation, demonstrating significantly better perplexity than an RNN trained by BPTT. Nevertheless these original RSM results are insightful regarding the behaviour of the algorithm.}
\begin{center}
\begin{tabular}{|c|c|c|c|}
\hline
\textbf{Model} & \textbf{Perplexity} \\
\hline
Random clusterings LM \cite{mikolov2011empirical} & 170 \\
\hline
\textbf{RSM} & \textbf{166} \\
\hline
5-gram Good-Turing \cite{mikolov2011empirical} & 162 \\
\hline
Structured LM \cite{mikolov2011empirical} & 146 \\
\hline
KN5 (our impl.) & 143 \\
\hline
Max. entropy 5-gram \cite{mikolov2011empirical} & 142 \\
\hline
Feed-forward ANN LM (deep BP) \cite{mikolov2011empirical} & 140 \\
\hline
Random forest LM \cite{mikolov2011empirical} & 131 \\
\hline
RNN LM w. BPTT \cite{mikolov2011empirical} & 124 \\
\hline
\textbf{RSM + KN5} & \textbf{124} \\
\hline
\textbf{bRSM} \cite{gordon2019learning} & \textbf{103} \\
\hline
LSTM variants \cite{merity2017regularizing} & $58 \leq 86$ \\
\hline
Transformer (GPT-2), varying size \cite{radford2019language} & $35 \leq 65$  \\
\hline
\end{tabular}
\end{center}
\label{table:ptb}
\end{table}

Table \ref{table:ptb} shows PTB test perplexity for RSM and selected popular language models. RSM is comparable to some early methods such as 5-gram Good-Turing, Random Clusterings, Random Forest and 5-gram Maximum Entropy, both individually and interpolated with KN5. The natural point of comparison is conventional RNNs trained with short-context BPTT. RNNs trained with BPTT are significantly better than RSM, with PPL 124 (RNN) and 105 (RNN+KN5). 

Carefully regularized, gated-memory ANN language models such as LSTM reduce PTB test perplexity from 124 to 100 or less \cite{zaremba2014recurrent}, with state of the art scores $< 60$ \cite{merity2017regularizing}. Models with attentional filtering are even better \cite{radford2019language}. However, these improvements required several years' effort by the research community, and exploit selective remembering features that could also be added to RSM over time.

Gordon et al \cite{gordon2019learning} recently published a modified version of RSM (bRSM) that replaces inhibition with a boosting scheme. The resultant PTB perplexity of 103 is significantly better than RNNs trained with BPTT (PPL 124) \cite{mikolov2011empirical}. This suggests that bRSM overcomes most or all of the generalization issues demonstrated by RSM. 

%To put these results in context, Mikolov et al \cite{mikolov2010recurrent} reports next-word perplexity using an Elman RNN, like RSM, but without sparse coding and inhibition to capture higher order sequences. They also interpolate with KN5. On the Wall St. Journal (WSJ) dataset they obtained 225 PPL when training on 1 million words (similar to PTB corpus size) and 156 when training with 6.4 million words.

\section{Summary}
\label{sec:summary}
Our primary objective was to develop a strictly local, biologically-plausible credit assignment strategy with comparable practical performance to early deep-backpropagation results. We believe we have been successful in this regard. On several tasks RSM provides results comparable to neural methods trained by deep BP or BPTT, which should not be possible for a simple RNN. For example, we showed that RSM can successfully learn to exploit distant causes when predicting later effects in the ERG task, which an RNN without BPTT cannot do, and in fact requires gated memory such as LSTM \cite{hochreiter1997long}. RSM can learn higher-order sequences, likely with combinatoric representational efficiency, despite some degree of sequential uncertainty and partial observability. On the ``bouncing balls'' task, RSM produced new best results beating a range of algorithms trained by deep BP, including LSTM. The learned model is also convincing in self-looped mode, generating arbitrarily long sequences of quite realistic billiard-ball dynamics.

RSM has a number of advantages - an order of magnitude reduction in memory requirements, very large memory capacity, batch-online training, and it is not necessary to specify the maximum time horizon in advance. For complex sequence learning tasks with similar training and test statistics, the RSM approach is appealing due to its simplicity and speed. However, although stochastic processes are not confounding, generalization to \emph{unseen} test sequences with different statistics is clearly limited in its current form, as seen in the PTB task. A later development of RSM by Gordon et al \cite{gordon2019learning} was able to overcome this limitation, giving better results on PTB than RNNs trained by BPTT. This reinforces the evidence that RSM represents a viable alternative credit assignment strategy to deep backpropagation through time. %In a language modelling task, when longer context is unnecessary, and training data is limited, this is a significant drawback.

In the image sequence prediction task, RSM was continually exposed to unseen sequences of observations due to the large number of images of each digit. Why was this form of generalization not pathological? First, the memorization approach is unlikely to confound spatial generalization; it is only unseen sequence generalization that is limited. Second, it is not possible to memorize the infinite observed training sequences, and the training regime therefore forces RSM to learn a generalized sequence model. Since both training and test set are drawn from the same distributions, this form of generalization is perhaps only interpolation rather than extrapolation.

%We suspect that this training regime forced RSM to preferentially use the available generalizable features of digits, whereas in language modelling all instances of a word-embedding are identical allowing RSM to overfit the observed word-sequences.

\section{Future work: Selective remembering}
%We expect that RSM would perform better as a language model given a larger training corpus, but would that merely be masking a fundamental weakness? 

One of the most significant differences between RSM and gated memory layers such as LSTM is the strategy of ``remember everything'' versus ``selectively remember useful things''. The latter approach is able to generalize better, because it is not distracted by irrelevant features in unseen test data: selective remembering can make different contexts become similar.

Selective remembering can also be considered a form of attention. Recent work has shown that self-attention is a very effective mechanism for selecting useful features from sequences \cite{Vaswani2017}, with groundbreaking results in neural language models. Radford et al \cite{radford2019language} use self-attention in an unsupervised next-word-prediction training regime, like RSM, but autoregressive, feed-forward, rather than recurrent. 

In future work we will attempt to use self-attention to selectively remember (sub) sequences using RSM, without violating our biological plausibility constraints. 

\section*{Acknowledgment}
The authors would like to thank Alan Zhang, for conversations about the nature of generalization and learning representations.

\bibliography{rsm.bib}
\bibliographystyle{IEEEtran}

\end{document}